\documentclass[10pt,twocolumn,letterpaper]{article}

\usepackage[pagenumbers]{cvpr} 

\definecolor{cvprblue}{rgb}{0.21,0.49,0.74}
\usepackage[pagebackref,breaklinks,colorlinks,allcolors=cvprblue]{hyperref}
\usepackage{xcolor}
\usepackage{algorithm}
\usepackage{algpseudocode}

\usepackage{booktabs}
\usepackage{multirow}
\usepackage{graphicx}
\usepackage[table]{xcolor}

\definecolor{proprietarybg}{HTML}{B7DED5}
\definecolor{opensourcebg}{HTML}{F0E2D6}
\definecolor{reasoningbg}{HTML}{D4CBE4}
\definecolor{ourmodelsbg}{HTML}{E99F72}
\definecolor{ourrowbg}{HTML}{DCEAFE}

\def\confName{CVPR}
\def\confYear{2027}

\title{From Reward Engineering to Data-Centric Reinforcement Learning for Effective Agentic Visual Tool Use}

\author{
    \textbf{Jin Cui}\textsuperscript{\rm 1 $\dagger$},
    \textbf{Xinyue Long}\textsuperscript{\rm 1,\rm 2 $\dagger$},
    \textbf{Jiepeng Zhou}\textsuperscript{\rm 3},
    \textbf{Boran Zhao}\textsuperscript{\rm 1,\rm 2*},
    \textbf{Pengju Ren}\textsuperscript{\rm 1},
    \textbf{Hao Dong}\textsuperscript{\rm 4} \vspace{4mm}\\
    \textsuperscript{\rm 1}State Key Laboratory of Human-Machine Hybrid Augmented Intelligence,\\
    and Institute of Artificial Intelligence and Robotics, Xi'an Jiaotong University\\
    \textsuperscript{\rm 2}School of Software Engineering, Xi'an Jiaotong University \\
    \textsuperscript{\rm 3}The Hong Kong University of Science and Technology (Guangzhou)\\
    \textsuperscript{\rm 4}ELLIS Institute Finland and Tampere University  
}

\begin{document}
\maketitle
{
    \renewcommand{\thefootnote}{\fnsymbol{footnote}}  
    \footnotetext[2]{Equal contribution} 
    \footnotetext[1]{Corresponding author.}
}
\begin{abstract}

Multimodal large language models (MLLMs) increasingly use external visual tools to acquire evidence during reasoning, yet learning when to invoke these tools remains challenging. Existing tool-use reinforcement learning (RL) methods often rely on call-contingent rewards to encourage beneficial tool use, but such reward engineering can distort the task objective and lead to degenerate calling behaviors. We argue that effective tool-use learning requires identifying not merely which questions are difficult, but which failures are attributable to missing visual evidence and can therefore be corrected through tool observations. We propose AOVO, a data-centric framework that enables MLLMs to learn effective tool use through perception attribution and policy-adaptive data sampling (PADS), without introducing tool-specific rewards. It counterfactually probes the model's reasoning states to identify observation-sensitive questions, distinguishing them from reasoning-dominated ones. PADS then periodically estimates category-level observation effects from recent policy rollouts and dynamically reallocates the training distribution toward questions where visual observations provide informative learning signals, while optimizing solely for final-answer correctness. Experiments across diverse visual perception and multimodal reasoning benchmarks demonstrate consistent improvements over competitive tool-use MLLMs, with causal analysis further validating the data-centric perspective for learning effective visual tool use.


\end{abstract}    
\section{Introduction}
\label{sec:intro}

Multimodal large language models (MLLMs) increasingly reason with images rather than merely about them: \textit{thinking-with-images} methods enable models to crop, zoom, or query external vision experts during reasoning to recover evidence missed by the initial visual encoding \citep{zheng2025deepeyes,hong2025deepeyesv2,wu2026vtoolr1,su2025openthinkimg,zhang2025chainoffocus}. Yet effective tool use requires knowing \emph{when} additional observations matter, which remains poorly calibrated. Recent audits show that many questions solved with tools are also solvable without them \citep{guo2026domm}, returned observations often leave answers unchanged \citep{wang2026illusion}, and tool-use RL improves largely by avoiding tool-induced harm rather than repairing intrinsic failures \citep{ma2026med}. Existing methods address this problem through call bonuses \citep{zheng2025deepeyes}, sample-level benefit weights \citep{wang2026adatooler}, judge-scored evidence paths \citep{long2026ntep}, or counterfactual call credit \citep{feng2026taco,peng2026countercredit}. These call-contingent objectives introduce additional incentives that can distort task optimization and induce either excessive or insufficient tool use \citep{wu2026vtoolr1,kim2026diversity,kang2026axpo}. We revisit tool-use learning from a data-centric perspective: \emph{rather than rewarding calls, we identify the questions that provide informative signals for learning when to observe.}

Our key insight is that \emph{\textbf{tool-use learning requires attributing outcome variation to observation rather than reasoning stochasticity}}. A failure is repairable only when missing visual evidence, rather than reasoning over available evidence, is the bottleneck; with heterogeneous experts, this effect is expert-specific. We identify it by branching from the learner's own reasoning states and comparing continuations without observation and with each expert, assigning questions to five categories: \emph{Repairable} (at least one expert improves answer success), \emph{Harmful} (expert observations reduce success), \emph{Reasoning-dominated} (outcomes vary across continuations but remain largely unaffected by observation), \emph{Saturated} (consistently solved), and \emph{Hard} (largely unsolved regardless of observation). These categories characterize the learner's response to observation rather than intrinsic question difficulty. We retain initial question assignments during RL but periodically re-estimate category-level observation effects under the evolving policy, enabling the training distribution to adapt without repeatedly reclassifying individual questions.


This distinction explains why tool-use RL remains unstable even with carefully designed rewards. On \textit{Reasoning-dominated} questions, outcomes vary across continuations but remain largely unaffected by observation; calls may be reinforced simply because they co-occur with successful reasoning, diluting observation-specific signals and destabilizing training, resulting in increased ineffective calls. We retain \textit{Hard} questions separately to assess whether alternative experts can recover them. Moreover, directly rewarding estimated repairability can drive near-never- or near-always-call regimes without improving call selection, while outcome-only RL can reinforce useful calls only when training exposes the learner's own repairable failures, which teacher demonstrations may not capture \citep{toolvision2026,openvistool2026}. These observations motivate a data-centric alternative: \emph{\textbf{rather than introducing increasingly elaborate tool rewards, identify and prioritize questions that provide observation-attributable learning signals}}. This perspective complements adaptive sampling based on difficulty, competence, or generic learning signals by prioritizing questions according to counterfactual observation effects \citep{zeng2026rlve,yu2025dapo,zhou2025revpt}.

We therefore propose \textbf{AOVO} (\emph{\underline{A}ttributing \underline{O}utcome \underline{V}ariation to \underline{O}bservation}), a data-centric framework for visual tool-use learning through learner-conditioned initialization and policy-adaptive data sampling rather than tool-contingent rewards. \emph{\textbf{Self-aware repair distillation}} first probes the learner's own reasoning states with each expert and converts successful repairs into supervision, establishing tool-use support where observation is useful to the learner itself. \emph{\textbf{Outcome-only GRPO}} optimizes the policy solely for final-answer correctness, without introducing call-dependent reward terms. During RL, \emph{\textbf{Policy-Adaptive Data Sampling (PADS)}} periodically re-estimates category-level observation effects under the current policy and dynamically routes training toward perception-dependent questions while retaining \textit{Reasoning-dominated} and \textit{Hard} examples as complementary training signals, so that sampling tracks the current perception boundary, thereby turning the outcome into a signal about looking rather than about reasoning. By reusing available trajectories and asynchronously generating only missing counterfactual branches, PADS incurs only 3.5\%--3.8\% additional rollout compute. Experiments across 13 benchmarks show consistent improvements over competitive tool-use models, with causal analysis further validating the call effectiveness. Our contributions are:


\begin{itemize}
    \item \textbf{A data-centric perspective on agentic visual tool use.} To our knowledge, we are the first to shift the focus from increasingly elaborate call-contingent rewards to identifying and prioritizing questions that provide observation-attributable learning signals.
    \item \textbf{AOVO with policy-adaptive data sampling.} The framework tracks category-level observation effects under the evolving policy and adaptively reallocates RL training data according to the current perception boundary. 
    \item \textbf{Comprehensive empirical experiments and causal validation.} Experiments across perception, reasoning, and general understanding benchmarks demonstrate leading performance on 12 accuracy-based benchmarks. Causal analyses further demonstrate consistent positive net gains.
\end{itemize}

\begin{figure*}
    \centering
    \includegraphics[width=1\linewidth]{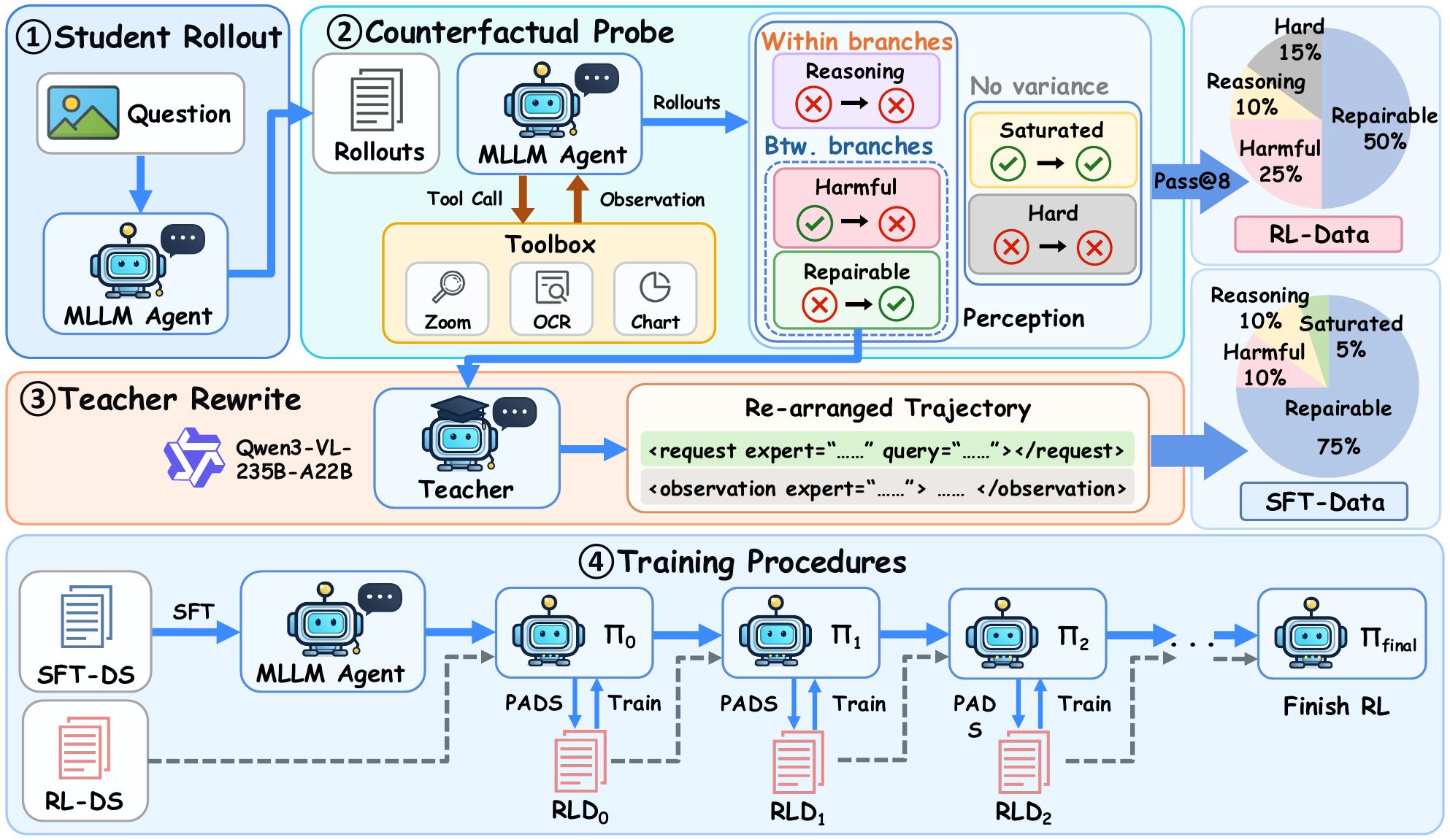}
    \caption{\textbf{Overview of AOVO.} (a)~\textbf{Student rollout}.(b)~\textbf{Counterfactual probe}: from the same reasoning state, the student continues both without and with each expert's observation. Comparing the outcomes assigns each question to one of five categories. (c)~\textbf{Teacher rewrite}: the teacher corrects only the reasoning after the observation in \textit{Repairable} trajectories, which then form the SFT data. (d)~\textbf{Training}: after SFT, the model is trained with outcome-only GRPO, and PADS periodically adjusts how often each category is sampled.}
    \label{fig:main-fig}
    \vspace{-4mm}
\end{figure*}

\section{Related Work}

\paragraph{Agentic Visual Tool Use in MLLMs.}
Recent work enables MLLMs to acquire visual evidence through cropping, zooming, code execution, and external vision experts \citep{zheng2025deepeyes,su2025openthinkimg,hong2025deepeyesv2,zhou2025revpt}. Most RL methods optimize a group-relative policy gradient with a composite trajectory reward:

\[ 
\label{eq:engineered}
R(\tau) = \underbrace{r(\tau)}_{\text{outcome}}
+ \lambda_{\mathrm{f}}\, r_{\mathrm{fmt}}(\tau)
+ \Big[\lambda_{\mathrm{t}}\, w(q)\, r_{\mathrm{tool}}(\tau) - \lambda_{\mathrm{n}}\, n(\tau)\Big]
\]
\vspace{-2mm}

where $r_{\mathrm{fmt}}$ enforces call and answer formats, $n(\tau)$ counts tool calls, and the bracketed terms are tool-contingent. Existing methods instantiate these terms through \emph{conditional bonuses}, rewarding correct tool-using trajectories \citep{zheng2025deepeyes}; \emph{necessity weighting}, scaling bonuses by offline-estimated per-question tool benefits, often with an overuse penalty \citep{wang2026adatooler,toolvision2026}; \emph{mode-adaptive rewards}, discounting tool-using rollouts when tool-free alternatives succeed within the same group \citep{beacon2026}; and \emph{process rewards}, evaluating calls against judge assessments or teacher-derived evidence paths \citep{long2026ntep}. \emph{Counterfactual credit} instead leaves $R$ unchanged and adjusts call-token advantages as $A_{i,t}\leftarrow A_i+\eta\,\kappa(c)$, where $\kappa(c)$ estimates call utility through branching or call replacement \citep{feng2026taco,peng2026countercredit}. Some methods use only outcome and format rewards after supervised cold-start \citep{hong2025deepeyesv2}.

These approaches shape tool-use behavior through reward design or credit assignment, yet improved accuracy does not necessarily imply effective tool use. MED \citep{ma2026med} finds that gains primarily arise from improved intrinsic capability and reduced tool-induced harm, with limited repair of intrinsic failures, while CauAudit \citep{wang2026illusion} reveals that many calls have little causal influence on final answers. Beyond call-contingent rewards or credit assignment, we use observation attribution to select informative training samples and optimize solely for outcome correctness.

\vspace{-2mm}

\paragraph{Adaptive Data Sampling.} Curriculum and self-paced learning adapt training examples to the learner's evolving competence \citep{bengio2009curriculum,kumar2010selfpaced}. Recent RL methods extend this idea through reward-variance filtering (DAPO) \citep{yu2025dapo}, adaptive sampling of informative prompts (Reinforce-Ada) \citep{xiong2025reinforceada}, and policy-adaptive difficulty scheduling (RLVE and ADS) \citep{zeng2026rlve,xu2026ads}. Unlike these methods, which prioritize samples based on difficulty, competence, or generic learning signals, we track a policy-dependent \emph{perception frontier}: whether expert observations change the task outcome. We periodically re-estimate this boundary and adjust sampling frequencies across the four question groups, prioritizing perception-dependent questions while revisiting the others as the policy evolves.

\newcommand{\tbd}[1]{\textcolor{red}{#1}}

\section{Attribute Outcome Variation to Observation}
\label{sec:method}

We first describe how we attribute outcome variation to observation and build the initial training data (Sec.~\ref{sec:partition}), how the training distribution tracks the moving perception boundary during RL (Sec.~\ref{sec:routing}), and the outcome-only objective that optimizes the policy on the routed data (Sec.~\ref{sec:grpo}). Fig.~\ref{fig:main-fig} gives an overview.


\vspace{-2mm}

\paragraph{Problem setup.}
Each question $q=(I,x,y^\star)$ consists of an image $I$, a text query $x$, and a verifiable answer $y^\star$. The policy $\pi_\theta$ has access to a set of vision experts $\mathcal{E}=\{e_1,\dots,e_K\}$ (\eg, zoom, OCR, chart parsing) as tools. A trajectory $\tau$ interleaves reasoning tokens with expert calls. A call is issued as \texttt{<request expert=$e$ query=$u$>}, and the environment returns an observation $o=e(I,u)$. The trajectory terminates with a predicted answer $\hat y(\tau)$. We denote a reasoning state, \ie, a prefix of a trajectory, by $s$, and answer correctness by $r(\tau)=\mathbb{1}[\hat y(\tau)=y^\star]$.

\subsection{Training Data Curation}
\label{sec:partition}

\paragraph{Counterfactual probe.}
Starting from student rollouts on the candidate dataset, we probe each reasoning state $s$ with every visual expert $e$. For each expert, the policy issues a request at $s$, and we sample $m$ continuations both with and without the returned observation:
\begin{equation}
\label{eq:probe}
\hat p^{\,e}(s)=\tfrac{1}{m}\textstyle\sum_{j=1}^{m}r(\tau_j^{\,s\oplus o_e}),\quad
\hat p^{\,\varnothing}(s)=\tfrac{1}{m}\textstyle\sum_{j=1}^{m}r(\tau_j^{\,s}).
\end{equation}
Here, $s$ is the prefix preceding the first call, or the final answer if the rollout makes no call. We use $m=4$ and temperature $1.0$. Let $\Delta_e(s)=\hat p^{\,e}(s)-\hat p^{\,\varnothing}(s)$, we attribute the outcome variation at $s$ either to observation (between branches) or to reasoning stochasticity (within branches). A state is \emph{Repairable} if some expert satisfies $\Delta_e(s)\ge\delta$, and \emph{Harmful} if no expert repairs it and the policy-selected expert yields $\Delta_e(s)\le-\delta$. Otherwise, it is \emph{Saturated} if all branches achieve success rates of at least $1-\epsilon$, \emph{Hard} if all are at most $\epsilon$, and \emph{Reasoning-dominated} otherwise. This separates observation-attributable outcome changes across branches from reasoning stochasticity within branches. We set $\delta=0.5$ and $\epsilon=1/m$. Hard and Reasoning-dominated states are merged as \emph{Reasoning} for SFT.

\paragraph{SFT initialization.}
We construct supervision according to the assigned categories (Fig.~\ref{fig:main-fig}). \emph{Repairable} states yield call trajectories preserving the student's reasoning prefix and expert request, followed by the returned observation and a correct continuation. For \emph{Harmful} and \emph{Saturated} states, we retain the student's correct tool-free continuations, while \emph{Hard} and \emph{Reasoning-dominated} states receive teacher-corrected tool-free supervision. 


\paragraph{Label validation.}
We validate assignments through three controls. (i) \emph{Observation literacy}: Limited ability to exploit expert observations may cause underestimation. After observation-conditioned warm-up, a student copy reproduces 94.1\% of assignments (Cohen's $\kappa=0.89$), with category proportions differing by at most 1.6\%. (ii) \emph{Sampling noise}: observation-free null probes yield only $2.3\%\pm0.5\%$ false \emph{Repairable}/\emph{Harmful} assignments, versus 53\% under actual observations, motivating $\delta=0.5$. (iii) \emph{Continuation faithfulness}: Qwen3-VL-235B-A22B verifies \emph{Repairable} trajectories and revises only the post-observation continuations that misinterpret or underuse the observation (approximately 12\%), preserving the original prefix and request.

\paragraph{RL initialization.}
For the remaining candidate questions, we apply pass@8 filtering using eight spontaneous rollouts from the SFT policy. Uniformly correct questions are assigned to \emph{Saturated}, while uniformly incorrect questions with no expert repair are assigned to \emph{Hard}; the remaining questions are categorized by counterfactual probing. This yields five fixed categories, with \emph{Saturated} examples excluded from RL and PADS adapting the sampling probabilities over the other four (Sec.~\ref{sec:routing}).

\subsection{Policy-Adaptive Data Sampling}
\label{sec:routing}

The initial categorization reflects the SFT policy, whereas the effect of visual observations evolves during RL. Reclassifying individual questions would require repeated counterfactual evaluations with limited statistical support. Instead, we retain the initial category assignments and adapt the sampling distribution using category-level estimates of observation effects under the evolving policy.

Let $\mathcal{D}_{\mathrm{gain}}$, $\mathcal{D}_{\mathrm{harm}}$, $\mathcal{D}_{\mathrm{reason}}$, $\mathcal{D}_{\mathrm{hard}}$, and $\mathcal{D}_{\mathrm{sat}}$ denote the five data subsets. At each RL step, questions are sampled according to category probabilities $w_k$, followed by uniform sampling within the selected subset. The Saturated subset is excluded, while the other remain eligible for training.

\paragraph{Counterfactual effect estimation.}
Every $T$ optimization steps, we estimate category-level observation effects from rollouts collected over the preceding $W$ steps, reusing factual outcomes and generating only missing counterfactuals. For called trajectories, we branch from the prefix preceding the first call with tool requests disabled; for tool-free trajectories, we introduce a call opportunity before the final answer, allowing the policy to select an expert and formulate its query. Both branches share the same reasoning prefix.

For category $k$, let $\mathcal{C}_k$ and $\mathcal{U}_k$ denote trajectories with and without expert calls, respectively. We estimate the average treatment effects on treated and untreated trajectories as

\vspace{-2mm}
\begin{equation} \small
\label{eq:category_effects}
\begin{aligned}
\mathrm{ATT}_k &=
\frac{1}{|\mathcal{C}_k|}
\sum_{i\in\mathcal{C}_k}
\left(y_i^{\mathrm{call}}-y_i^{\varnothing}\right),\\
\mathrm{ATU}_k &=
\frac{1}{|\mathcal{U}_k|}
\sum_{i\in\mathcal{U}_k}
\left(y_i^{\mathrm{call}}-y_i^{\varnothing}\right),
\end{aligned}
\end{equation}
\vspace{-2mm}

where $y^{\mathrm{call}},y^{\varnothing}\in\{0,1\}$ indicate answer correctness with and without observation. ATT captures the effect of actual calls, while ATU estimates their effect at abstained states. Let $c_k$ denote the category-level calling rate.

We define the sampling priority of category $k$ as

\vspace{-4mm}
\begin{equation} \small
\label{eq:category_score}
s_k ={}
\underbrace{c_k[\mathrm{ATT}_k]_+}_{\text{observed call benefit}}
+ \underbrace{(1-c_k)[\mathrm{ATU}_k]_+}_{\text{missed potential benefit}}
+ \underbrace{c_k[-\mathrm{ATT}_k]_+}_{\text{observed call harm}},
\end{equation}

where $[x]_+=\max(x,0)$. The three terms capture realized call benefits, missed opportunities, and call-induced harm, respectively. Harmful calls remain informative for learning when to abstain.

For Hard questions, we additionally probe failed call trajectories with an alternative expert. Let $\mathrm{ALT}_{\mathrm{hard}}$ denote the success rate of alternative-expert continuations conditioned on original-call failure. We augment the score by

\vspace{-2mm}
\begin{equation}
\label{eq:hard_score}
s_{\mathrm{hard}} \mathrel{+}=
c_{\mathrm{hard}}
(1-\bar y_{\mathrm{hard}}^{\mathrm{call}})
\mathrm{ALT}_{\mathrm{hard}},
\end{equation}
\vspace{-4mm}

where $\bar y_{\mathrm{hard}}^{\mathrm{call}}$ is the mean correctness of original call trajectories, capturing additional repair opportunities through expert reselection.

\paragraph{Adaptive sampling distribution.}
We normalize category scores and update sampling probabilities via an exponential moving average:

\vspace{-4mm}
\begin{equation}
\label{eq:sampling_update}
\tilde w_k = \frac{s_k}{\sum_{j\in\mathcal{K}}s_j}, \quad
w_k \leftarrow (1-\alpha)w_k+\alpha\tilde w_k,
\end{equation}
\vspace{-4mm}

where $\mathcal{K}=\{\mathrm{gain},\mathrm{harm},\mathrm{reason},\mathrm{hard}\}$ and $\alpha=0.5$. Updates are skipped if $\sum_j s_j<0.01$.

To maintain coverage of all active categories, we enforce minimum sampling probabilities of $(0.40,0.20,0.10,0.05)$ for \textit{Repairable}, \textit{Harmful}, \textit{Reasoning-dominated}, and \textit{Hard} examples, respectively, and redistribute the remaining probability mass proportionally. These constraints preserve exposure to successful repairs, harmful calls, general reasoning examples, and unresolved questions with potential expert-assisted solutions. Sampling starts from $w^{(0)}=(0.50,0.25,0.10,0.15)$, with $T=100$ and $W=15$.

\paragraph{Efficient online probing.}
We reduce probing overhead by reusing stored reasoning prefixes and expert observations, limiting sampling to two trajectories per question and restricting evaluation to recent rollouts. Categories with fewer than 32 distinct questions are supplemented with probe-only rollouts. Counterfactual continuations are excluded from GRPO and used solely for sampling updates. Compared with DAPO's rejection-based dynamic sampling, which incurs an expected oversampling overhead of $r/(1-r)$ for rejection ratio $r$, our approach adds only $3.5\%$--$3.8\%$ rollout compute, equivalent to a rejection rate of $3.4\%$--$3.7\%$ under equal per-group generation costs \citep{yu2025dapo}.

\subsection{Outcome-Only Policy Optimization}
\label{sec:grpo}

We optimize the policy with GRPO~\cite{shao2024deepseekmath} on the adaptive sampling distribution $\mathcal{D}_t$ induced by PADS, using final-answer correctness
as the sole reward:
\begin{equation}
\label{eq:reward}
R(\tau)=r(\tau)=\mathbb{1}[\hat y(\tau)=y^\star].
\end{equation}
Trajectories that yield no valid, parseable answer receive zero reward, eliminating the need for a separate format term. For each $q\sim\mathcal{D}_t$, we sample $G$ trajectories $\{\tau_i\}_{i=1}^G$ from $\pi_{\theta_{\mathrm{old}}}$ and compute group-normalized advantages:
\begin{equation} \small
A_i=
\frac{R(\tau_i)-\operatorname{mean}_j R(\tau_j)}
{\operatorname{std}_j R(\tau_j)+\epsilon_{\mathrm{norm}}}.
\end{equation}
We then maximize

\vspace{-4mm}
\begin{equation} \small
\label{eq:grpo}
\begin{aligned}
\mathcal{J}(\theta)
=&\ \mathbb{E}\Bigg[
\frac{1}{\sum_i|\mathcal{M}_i|}
\sum_{i=1}^{G}\sum_{t\in\mathcal{M}_i}
\min\Big(
\rho_{i,t}A_i,\,\\
&
\operatorname{clip}(\rho_{i,t},1-\varepsilon,1+\varepsilon)A_i
\Big)\Bigg]-\beta\,\mathbb{D}_{\mathrm{KL}}
\big(\pi_\theta\,\|\,\pi_{\mathrm{ref}}\big),
\end{aligned}
\end{equation}
\vspace{-2mm}

where $\rho_{i,t}=\pi_\theta(\tau_{i,t}\mid\tau_{i,<t})/
\pi_{\theta_{\mathrm{old}}}(\tau_{i,t}\mid\tau_{i,<t})$,
and $\mathcal{M}_i$ indexes the policy-generated tokens of $\tau_i$;
expert observations are excluded from the policy loss.

Under this objective, the expected return difference between calling and abstaining at a reasoning state $s$ is
\begin{equation}
\begin{aligned}
&\Delta_{\mathrm{call}}(s)
=\mathbb{E}[R\mid s,\mathrm{call}]
-\mathbb{E}[R\mid s,\mathrm{abstain}]\\
&=P(\mathrm{correct}\mid s,\mathrm{call})
-P(\mathrm{correct}\mid s,\mathrm{abstain}).
\end{aligned}
\end{equation}


Dynamic sampling therefore determines which questions drive learning, while outcome-only optimization determines which behaviors are reinforced. The complete training procedure is summarized in the Appendix.

\begin{table*}[t]
\centering
\caption{\textbf{Main results} on perception, reasoning, and general
multimodal benchmarks with Qwen2.5-VL-7B-Instruct as the base model. All entries are accuracy (\%), except MME (total score). Within the tool-use groups (explicit, implicit, and ours), the best result per column is in \textbf{bold} and the second best is \underline{underlined}. The last row reports the gain of AOVO over vanilla GRPO trained on the same data.}
\vspace{-2mm}
\label{tab:main-results}

\small
\setlength{\tabcolsep}{4pt}
\renewcommand{\arraystretch}{0.92}

\resizebox{\textwidth}{!}{%
\begin{tabular}{lccccccccccccc}
\toprule
\multirow{2}{*}{Model}
& \multicolumn{6}{c}{Perception}
& \multicolumn{3}{c}{Reasoning}
& \multicolumn{3}{c}{General}
& \multirow{2}{*}{POPE} \\
\cmidrule(lr){2-7}
\cmidrule(lr){8-10}
\cmidrule(lr){11-13}
& DocVQA & OCRBench & InfoVQA & HR-4K & HR-8K & V$^*$
& MVision & MVista & MVerse
& GQA & MMBench & MME
& \\
\midrule

\multicolumn{14}{c}{\textit{Proprietary models}} \\
\cmidrule(lr){1-14}
GPT-4o
& 92.8 & 73.6 & -- & 62.4 & 58.9 & 62.3
& 43.8 & 71.6 & 49.9
& 83.3 & 86.0 & 2329 \\
Gemini-2.5-Pro
& 92.6 & 86.6 & -- & 83.9 & 81.5 & 79.1
& 39.8 & 80.9 & 76.9
& -- & 88.3 & -- \\

\midrule
\multicolumn{14}{c}{\textit{Open-source MLLMs (no visual tools)}} \\
\cmidrule(lr){1-14}
Qwen2.5-VL-7B           
& 87.6 & 69.7 & 72.3 & 72.5 & 68.5 & 78.5  
& 22.7 & 68.2 & 35.2 
& 51.5 & 86.5 & 2309 
& 87.2 \\
Qwen3-VL-8B             
& 88.8 & 72.7 & 75.6 & 76.2 & 72.4 & 82.2 
& 23.0 & 77.2 & 62.1 
& 53.5 & 87.3 & 2573 
& 89.6 \\
InternVL3-8B            
& 85.6 & 66.4 & 72.3 & 71.0 & 62.0 & 74.9 
& 24.2 & 70.4 & 29.2 
& 53.2 & 87.5 & 2495 
& 90.3 \\

\midrule
\multicolumn{14}{c}{\textit{Explicit tool-use models (each with its distinct tools)}} \\
\cmidrule(lr){1-14}
Pixel-Reasoner-7B
& 86.9 & 77.5 & 72.0 & 73.0 & 63.6 & 84.3
& 26.9 & 71.2 & 45.9
& 50.1 & 88.2 & 2373 
& 85.5 \\
DeepEyes-7B
& 86.1 & 79.8 & 75.0 & 73.9 & 69.8 &  84.3
& \underline{27.9} & 70.1 & 42.1
& 53.2 & 86.0 & 2375 
& 88.0 \\
Mini-o3-7B
& 83.8 & 78.9 & 69.2 & 75.5 & 68.5 & \textbf{88.2}
& 21.1 & 67.2 & 45.4
& 55.2 & 85.5 & 2336 
& 88.1 \\
Thyme-7B
& 83.0 & 86.3 & 68.0 & 75.0 & 71.8 & 82.2
& 23.0 & 70.0 & 39.1
& 55.2 & \underline{89.6} & 2341 
& 85.6 \\
DeepEyesV2-7B
& -- & \underline{88.2} & -- & \underline{76.3} & 73.8 & 81.8
& \textbf{28.9} & \underline{71.8} & \textbf{52.7}
& 53.2 & 87.5 & 2375 
& -- \\
CodeV-RL-7B
& -- & 85.6 & -- & 76.1 & 71.3 & 84.8
& -- & 71.8 & \underline{49.2}
& 53.2 & 87.6 & 2235 
& \underline{88.7} \\

\midrule
\multicolumn{14}{c}{\textit{Implicit tool augmented models}} \\
\cmidrule(lr){1-14}
ZwZ-7B 
& 86.4 & 64.6 & 63.8 & 75.4 & 72.9 & \underline{88.0}
& 25.0 & 70.0 & 45.0 
& 49.5 & 88.3 & 2199 
& 88.6 \\
Imagine-OPD-8B       
& \underline{89.1} & 76.1 & 74.0 & 74.2 & \underline{74.4} & 86.9 
& 26.8 & 63.7 & 41.2 
& 48.2 & 87.9 & 2420 
& 87.5 \\
Vision-OPD-4B       
& 89.0 & 80.8 & \underline{76.8} & 64.1 & 59.6 & 62.8 
& 27.6 & 63.7 & 41.2 
& \underline{53.8} & 78.8 & \underline{2470}
& 86.7 \\

\midrule

\textbf{Vanilla GRPO} 
& 89.3 & 81.4 & 75.3 & 73.8 & 71.4 & 81.5 
& 20.7 & 67.0 & 40.7 
& 53.2 & 87.6 & 2406 
& 88.3 \\

\rowcolor{ourrowbg}
\textbf{Ours-7B}
& \textbf{93.2} & \textbf{88.3} & \textbf{77.9} & \textbf{77.0} & \textbf{74.9} & 84.3
& 25.3 & \textbf{71.9} & 42.7
& \textbf{57.2} & \textbf{90.0} & \textbf{2550} 
& \textbf{89.3} \\


$\triangle$ (\textit{vs} Vanilla GRPO) 
& +3.9 & +6.9 & +2.6 & +3.2 & +3.5 & +2.8 
& +4.6 & +4.9 & +2.0 
& +4.0 & +2.4 & +144 
& +1.0 \\

\bottomrule
\end{tabular}%
}
\end{table*}


\begin{figure*}
    \centering
    \vspace{-4mm}
    \includegraphics[width=1\linewidth]{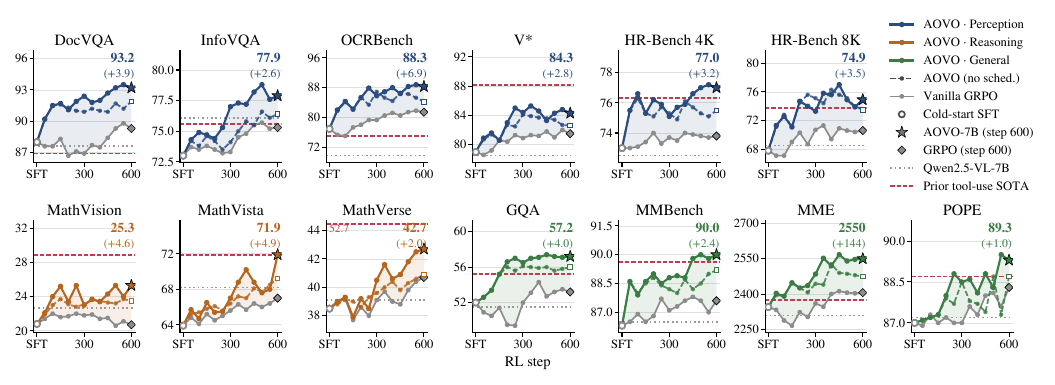}
    \vspace{-8mm}
    \caption{\textbf{Benchmark performance during RL.} Accuracy from cold-start SFT (open circles) to RL step 600. Solid blue, orange, and green lines denote AOVO on perception, reasoning, and general benchmarks; dashed lines denote AOVO without scheduling (fixed $w^{(0)}$), and gray lines denote vanilla GRPO. Shading indicates the AOVO--GRPO gap. Stars and diamonds mark final AOVO and GRPO checkpoints. Dotted and red dashed lines indicate Qwen2.5-VL-7B and prior tool-use SOTA for each benchmark, respectively.}
    \label{fig:accuracy_dynamics}
\end{figure*}


\section{Experiments}
\label{sec:exp}

\subsection{Experimental Setup}

\paragraph{Benchmarks.}
We evaluate our models on four complementary benchmark groups. \emph{Perception} includes DocVQA \citep{mathew2021docvqa}, OCRBench \citep{liu2024ocrbench}, and InfographicVQA \citep{mathew2022infographicvqa}, covering graphical and structural information from visual documents. \emph{Reasoning} comprises MathVerse \citep{zhang2024mathverse}, MathVista \citep{lu2024mathvista}, and MathVision \citep{wang2024mathvision} for challenging logical and visual mathematical reasoning. \emph{General} includes GQA \citep{hudson2019gqa} for compositional and relational visual understanding, alongside MMBench \citep{liu2024mmbench} and MME \citep{fu2023mme} for comprehensive multimodal evaluation. Finally, we include V* \citep{wu2024vstar} and HR-Bench at 4K and 8K resolutions \citep{wang2025hrbench} to further assess fine-grained perception under high-resolution settings.

\begin{figure*}
    \centering
    \includegraphics[width=1\linewidth]{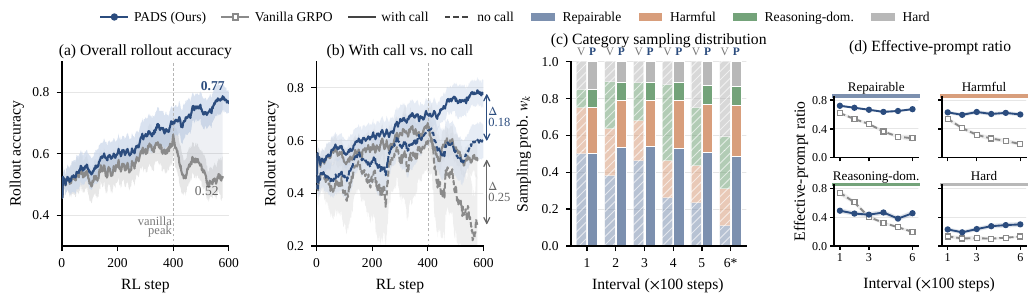}
    \vspace{-8mm}
    \caption{\textbf{Training dynamics of PADS vs.\ vanilla GRPO.} (a) Rollout accuracy, showing PADS's steady improvement versus GRPO's late collapse. (b) Accuracy with (solid) and without (dashed) expert calls; $\Delta$ denotes their final gap. (c) Category composition per 100-step interval: P denotes PADS sampling probabilities $w_k$, while V (hatched) denotes vanilla GRPO composition re-probed under its current policy ($6^*$: final probe at step 600). (d) Per-category effective-prompt ratio, indicating the fraction of groups with nonzero advantage.}
    \label{fig:dynamic_compare}
\end{figure*} 


\begin{figure*}
    \centering
    \includegraphics[width=1\linewidth]{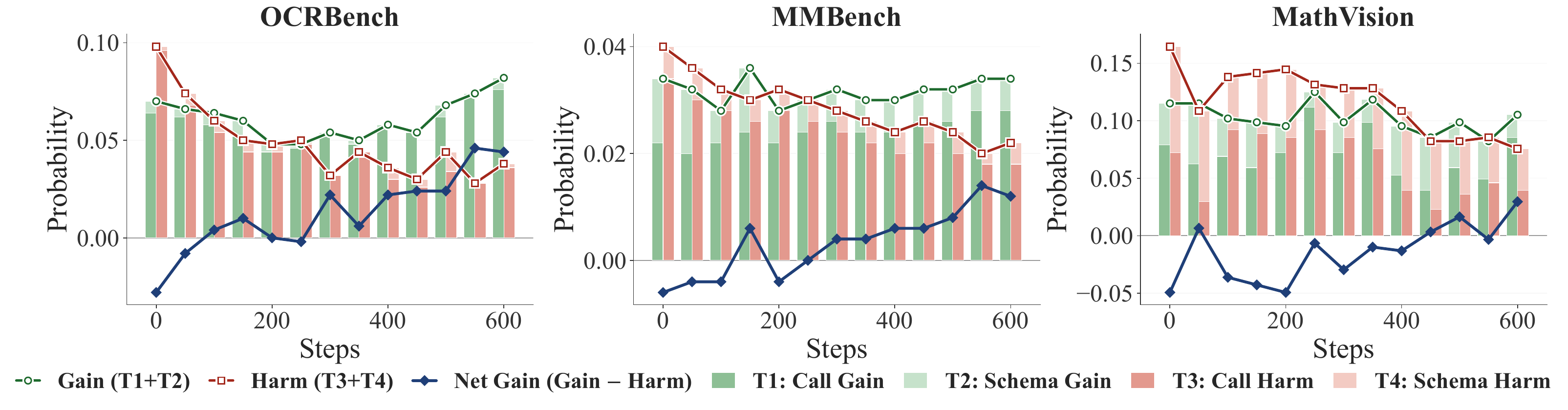}
    \vspace{-8mm}
    \caption{\textbf{Causal decomposition of tool-induced effects} following MED~\cite{ma2026med} on OCRBench, MMBench, and MathVision (perception, general, and reasoning). At each RL checkpoint, the tool-available vs. tool-free accuracy gap is decomposed into Call/Schema Gain (T1/T2) and Call/Schema Harm (T3/T4). Curves show total Gain (T1+T2), Harm (T3+T4), and Net Gain (Gain$-$Harm).}
    \label{fig:med_analysis}
\end{figure*}



\paragraph{Baselines.}
We compare against four groups of baselines. \emph{Proprietary models} include GPT-4o and Gemini-2.5-Pro \citep{openai2024gpt4o, comanici2025gemini25}. \emph{Open-source MLLMs} Qwen2.5-VL-7B \citep{bai2025qwen25vl}, Qwen3-VL-8B \citep{bai2025qwen3vl}, and InternVL3-8B \citep{zhu2025internvl3}. For \emph{explicit tool-use models}, we compare against Pixel-Reasoner-7B, Thyme-7B, Mini-o3-7B, CodeV-7B, DeepEyes-7B, and DeepEyesV2-7B \citep{wang2025pixelreasoner,zhang2026thyme,lai2026minio3,hou2026codev,zheng2025deepeyes,hong2025deepeyesv2}. We also include methods that internalize inspection without inference-time tool calls (\emph{implicit tool augmented models}), namely Vision-OPD-4B, ZwZ-7B, and Imagine-OPD-8B \citep{yuan2026visionopd,wei2026zooming,cai2026imagineopd}.



\paragraph{Data Curation.}
Our SFT dataset comprises 10K examples from GQA \citep{hudson2019gqa}, ChartQA \citep{masry2022chartqa}, TextVQA \citep{singh2019textvqa}, RobuT-WTQ \citep{zhao2023robut}, UReader-IE \citep{ye2023ureader}, and A-OKVQA \citep{schwenk2022aokvqa}, following the data composition in Fig.~\ref{fig:main-fig}. This data mixture balances general and structured visual perception and multimodal reasoning while ensuring broad task coverage. The RL dataset comprises 20K examples from GQA, RobuT-WTQ, and UReader-IE retained after pass@8 filtering, with 50\% \textit{Repairable}, 25\% \textit{Harmful}, 10\% \textit{Reasoning-dominated}, and 15\% \textit{Hard}; \textit{Saturated} questions are discarded for training efficiency. For OCR-related samples, we conduct answer verifiability and formatting constraints. To prevent evaluation leakage, we apply a decontamination following Rufus-Air \citep{chang2026rufusair} when constructing the train/test splits.




\paragraph{Training Details.}
We initialize our models from Qwen2.5 -VL-7B. For cold-start SFT, we train for 3 epochs, using AdamW with an initial learning rate of $1\times10^{-5}$. For RL, we perform GRPO \citep{shao2024deepseekmath} for up to 600 steps, with a learning rate of $2\times10^{-6}$ and a KL coefficient of 0.01. Each update consists of 32 prompts with 8 rollouts per prompt, yielding 256 trajectories per update. Our toolbox comprises three experts: GOT-OCR 2.0 \cite{wei2024general} for OCR, DePlot \cite{liu2023deplot} for chart parsing, and Grounding DINO \cite{liu2024grounding} with cropping for visual localization. We further validate the framework's flexibility by extending it to additional tools in the Appendix. Training and experiments are conducted on three nodes, each equipped with four NVIDIA GH200 GPUs.




\subsection{Main Results}

\paragraph{Perception and general multimodal understanding.}
AOVO achieves the best results within the tool-use groups on five of six perception benchmarks (Tab.~\ref{tab:main-results}), averaging 82.6 and surpassing DeepEyes (78.2) and Imagine-OPD (79.1) by 4.4 and 3.5 points, respectively. The only exception is V$^*$, where models specialized for iterative visual search (Mini-o3, ZwZ) remain ahead. Gains are largest on text-dense documents (+6.9 on OCRBench and +3.9 on DocVQA over vanilla GRPO), where missing evidence can be recovered through a single expert observation. Beyond perception, AOVO outperforms all tool-use methods on all four general multimodal benchmarks. This suggests that learning when to look transfers beyond training domains, improving general understanding while preserving tool-free competence, consistent with fewer hallucinated objects on POPE \citep{li2023pope} and the causal analysis in Fig.~\ref{fig:med_analysis}.

\vspace{-4mm}

\paragraph{Reasoning.}
AOVO improves over vanilla GRPO on all three math benchmarks (+4.6, +4.9, and +2.0 on MathVision, MathVista, and MathVerse) and achieves the best result on MathVista, but still trails the strongest explicit tool-use models on MathVision and MathVerse. This reflects PADS down-weighting \textit{Reasoning-dominated} and \textit{Hard} questions, whose outcomes are not attributable to observation: increasing their share improves reasoning only at the cost of perception and net gain (Tab.~\ref{tab:mix_sens}). We attribute the observed gains primarily to improved visual grounding rather than substantial improvements in reasoning; disentangling the two causally remains beyond our scope.

\paragraph{Causal analysis of tool use.}
Following MED~\cite{ma2026med}, we decompose the tool-induced accuracy gap into Call Gain (T1), Schema Gain (T2), Call Harm (T3), and Schema Harm (T4). MED finds that tool use accounts for only 22--30\% of the improvement in crop-and-zoom RL, where gains primarily come from reduced harm while Call Gain stagnates. AOVO breaks this pattern despite using heterogeneous experts (Fig.~\ref{fig:med_analysis}): on OCRBench, Call Harm declines while Call Gain rises, increasing net gain from $-2.8$ after SFT to $+4.4$; MMBench likewise turns positive, while MathVision remains near zero, consistent with the reasoning results. This difference stems from data rather than reward design. Engineered call rewards favor harm reduction by incentivizing the suppression of risky calls, whereas AOVO's net gain depends on data composition (Tab.~\ref{tab:mix_sens}): removing \textit{Harmful} data or oversampling \textit{Repairable} or \textit{Reasoning-dominated} data drives net gain to $-3.5$, $-4.3$, and $-3.1$, respectively. Separating perception-dependent from reasoning-dominated questions thus suppresses spurious calls while retaining those that repair failures.


\section{Analysis}

\begin{table}[t]
\centering
\scriptsize
\setlength{\tabcolsep}{1pt}
\renewcommand{\arraystretch}{0.95}

\caption{\textbf{Sensitivity to the initial data mixture.}
Each experiment scales one category and renormalizes the others (excluding \textit{Saturated}), varying the three categories shared by SFT and RL. \emph{SFT stage}: vary SFT proportions with default RL initialization $w^{(0)}$; \emph{RL stage}: vary $w^{(0)}$ with SFT and PADS fixed. Reference rows report the base model, cold-start SFT, and vanilla GRPO (vGRPO). \emph{Share}: category proportion (\%); \emph{NG}: OCRBench net gain (Gain$-$Harm, Fig.~\ref{fig:med_analysis}, $\times100$), omitted for Call $<10\%$; \emph{Call}: test-time call rate (\%). Shading indicates significant deviations from the default (\colorbox{red!10}{over-calling}, \colorbox{blue!10}{under-calling}). Results use Qwen2.5-VL-7B, averaged over three seeds.}

\label{tab:mix_sens}

\resizebox{\columnwidth}{!}{%
\begin{tabular}{@{}l@{\hspace{0pt}}l@{\hspace{0.5pt}}ccccc@{\hspace{2pt}}ccccc@{}}
\toprule
& & \multicolumn{5}{c}{SFT-stage mixture}
& \multicolumn{5}{c}{RL-stage initial mixture $w^{(0)}$} \\
\cmidrule(lr){3-7}\cmidrule(lr){8-12}
Category & Scale & Share & Perc. & Reas. & NG & Call
& Share & Perc. & Reas. & NG & Call \\
\midrule

\multicolumn{2}{@{}l}{Base model}
& -- & 74.9 & 42.0 & -- & --
& -- & 74.9 & 42.0 & -- & -- \\

\multicolumn{2}{@{}l}{SFT\,/\,vGRPO}
& -- & 76.2 & 41.0 & -2.8 & 57.6
& -- & 78.8 & 42.8 & 0.6 & 31.0 \\
\cmidrule(lr){1-12}

\rowcolor{gray!12}
\multicolumn{2}{@{}l}{Default}
& -- & \textbf{82.6} & 46.6 & \textbf{4.4} & 45.2
& -- & 82.6 & 46.6 & \textbf{4.4} & 45.2 \\
\midrule

\multirow{4}{*}{Repairable}
 & $\times 0$        & 0.0  & 78.6 & 45.4 & -- & \cellcolor{blue!10}3.8
                     & 0.0  & 76.6 & 41.3 & -- & \cellcolor{blue!10}1.2 \\
 & $\times \tfrac12$ & 65.2 & 81.2 & 46.0 & 2.9 & 42.4
                     & 33.3 & 81.8 & 46.5 & 3.0 & 40.8 \\
 & $\times 2$        & 88.2 & 81.9 & 44.6 & 3.1 & 58.3
                     & 66.7 & \textbf{82.8} & 46.6 & 3.6 & 61.3 \\
 & $\times 4$        & 93.8 & 79.7 & 43.2 & -1.2 & \cellcolor{red!10}79.4
                     & 80.0 & 81.2 & 40.6 & -4.3 & \cellcolor{red!10}98.1 \\
\midrule

\multirow{4}{*}{Harmful}
 & $\times 0$        & 0.0  & 79.6 & 43.8 & -0.5 & 57.0
                     & 0.0  & 80.2 & 44.5 & -3.5 & \cellcolor{red!10}83.4 \\
 & $\times \tfrac12$ & 5.6  & 82.5 & 46.5 & 3.2 & 52.1
                     & 14.3 & 82.5 & 45.6 & 3.9 & 48.1 \\
 & $\times 2$        & 19.0 & 82.1 & 46.5 & 2.9 & 39.4
                     & 40.0 & 80.7 & 45.7 & 4.2 & 41.2 \\
 & $\times 4$        & 32.0 & 81.0 & 46.2 & 1.0 & 35.3
                     & 57.1 & 77.3 & 42.5 & -- & \cellcolor{blue!10}6.9 \\
\midrule

\multirow{4}{*}{Reasoning}
 & $\times 0$        & 0.0  & 82.1 & 44.3 & 2.1 & 47.0
                     & 0.0  & 82.3 & 42.5 & 1.6 & 42.1 \\
 & $\times \tfrac12$ & 5.6  & 82.4 & 45.9 & 3.0 & 46.2
                     & 5.3  & 82.0 & 45.8 & 4.2 & 45.0 \\
 & $\times 2$        & 19.0 & 82.3 & 46.8 & -0.2 & 49.3
                     & 18.2 & 81.2 & \textbf{46.7} & -0.5 & 52.4 \\
 & $\times 4$        & 32.0 & 80.2 & \textbf{46.9} & -2.3 & 58.9
                     & 30.8 & 78.9 & 46.0 & -3.1 & \cellcolor{red!10}72.8 \\
\bottomrule
\end{tabular}%
}
\end{table}


\paragraph{Training dynamics.}
Fig.~\ref{fig:dynamic_compare} contrasts PADS with vanilla GRPO. Vanilla GRPO peaks at 0.66 around step 400 before collapsing to 0.52, whereas PADS steadily reaches 0.77 (a). This coincides with a loss of informative signals: the re-probed pool shifts from 50\% \textit{Repairable} to 11\%, while \textit{Hard} and \textit{Reasoning-dominated} grow to 41\% and 28\% (c); effective-prompt ratios also decline across categories (e.g., 0.62 to 0.27 for \textit{Repairable}). Although \textit{Reasoning-dominated} initially has a high ratio (0.74), its advantages reflect reasoning stochasticity rather than observation effects, potentially reinforcing spurious calls. The resulting scarcity of informative groups destabilizes optimization. In contrast, PADS maintains \textit{Repairable} and \textit{Harmful} shares at 49--54\% and 25--28\%, respectively, while keeping \textit{Reasoning-dominated} at its 10\% floor, sustaining effective-prompt ratios of 0.65 and 0.61 for the former two categories. The ratio on \textit{Hard} also rises from 0.23 to 0.30, versus 0.13 under vanilla GRPO, suggesting that some initially unsolvable questions become recoverable through observation as the policy improves. Consequently, PADS reaches 0.78 and 0.60 accuracy with and without calls, respectively, while both vanilla GRPO branches deteriorate after step 400, particularly the no-call branch (0.59 to 0.29, versus 0.64 to 0.53 for the call branch) (b).

\vspace{-4mm}

\paragraph{Sensitivity to the initial data mixture.}
Tab.~\ref{tab:mix_sens} scales one category at a time.
\emph{(i) The RL mixture matters more.} Perturbing $w^{(0)}$ shifts accuracy, net gain, and call rate by up to 6.0, 8.7, and 52.9 points, respectively, versus 4.0, 6.7, and 41.4 for the SFT mixture. PADS largely absorbs SFT-stage imbalances, whereas a skewed $w^{(0)}$ can induce persistent all-call or no-call regimes before adaptation, as group advantages cease to contrast calling with abstaining. Removing \textit{Repairable} data from SFT is an exception: the call rate falls to 3.8\%, and RL gains only 0.6 points over vanilla GRPO. \emph{(ii) Categories play distinct roles.} Excess \textit{Repairable} data induces 98.1\% calling and the lowest net gain ($-4.3$). \textit{Harmful} data regulates abstention: excess suppresses calling to 6.9\%, while removal causes over-calling (83.4\%, net gain $-3.5$). \textit{Reasoning} data anchors general reasoning (removal costs 4.1 points), but excess triggers calls on failures that observations cannot repair (net gain $-3.1$), an effect not offset by \textit{Harmful} data. This supports separating the two categories. \emph{(iii) Moderate deviations are benign.} Scaling by $\times\tfrac12$ or $\times 2$ keeps accuracy within 2.0 points, indicating that AOVO requires no fine-grained mixture tuning. We vary one category at a time; joint exploration under stricter controls remains future work.

\begin{table}[t]
\centering\small
\setlength{\tabcolsep}{2.8pt}
\caption{\textbf{Sensitivity to the PADS re-probing interval $T$} ($W\!=\!15$). Cost: extra rollout compute (\%).
Perc.: mean$_{\pm\text{std}}$ over 3 seeds. Roll.: training rollout accuracy at step 600;
Drop: peak-to-final rollout-accuracy drop. $T\!=\!\infty$ keeps $w^{(0)}$ fixed.}
\label{tab:interval_sens}
\begin{tabular}{l c c c c c c c}
\toprule
$T$ & Cost & Perc. & Reas. & NG & Call & Roll. & Drop$\downarrow$ \\
\midrule
10       & 36.0 & 80.2$_{\pm1.4}$ & 45.4 & 2.9 & 51.8 & .68 & .09 \\
25       & 14.4 & 81.3$_{\pm0.9}$ & 46.0 & 3.6 & 48.6 & .72 & .05 \\
50       & 7.2  & 82.1$_{\pm0.5}$ & 46.4 & 4.1 & 46.3 & .75 & .02 \\
\rowcolor{gray!12}
100      & 3.6  & \textbf{82.6}$_{\pm0.2}$ & \textbf{46.6} & \textbf{4.4} & 45.2 & \textbf{.77} & \textbf{.02} \\
200      & 1.8  & 82.0$_{\pm0.3}$ & 46.3 & 4.0 & 43.1 & .73 & .02 \\
400      & 0.9  & 81.3$_{\pm0.3}$ & 45.9 & 3.5 & 40.6 & .69 & .03 \\
$\infty$ & 0.0  & 80.6$_{\pm0.3}$ & 44.5 & 3.3 & 39.4 & .67 & .04 \\
\midrule
Vanilla GRPO & 0.0 & 78.8$_{\pm0.6}$ & 42.8 & 0.6 & 31.0 & .52 & .14 \\
\bottomrule
\end{tabular}
\end{table}

\paragraph{Sensitivity to the re-probing interval.}
Tab.~\ref{tab:interval_sens} shows a trade-off in $T$. Frequent re-probing ($T\leq25$) incurs 14.4--36.0\% additional compute and destabilizes training (perception-score std up to 1.4; peak-to-final rollout drop of 0.09), as the EMA tracks noisy, correlated estimates. Conversely, infrequent re-probing allows $w$ to lag behind the moving perception boundary; at $T=400$, performance approaches the fixed-$w^{(0)}$ variant ($T=\infty$), which nevertheless outperforms vanilla GRPO through attribution-based curation alone. All $T\in[50,200]$ remain within 0.6 points of the best result, with $T=100$ achieving the best performance at only 3.6\% additional rollout compute.

\section{Conclusion}
We demonstrate the effectiveness of a data-centric perspective to RL-based agentic visual tool use, shifting the focus from call-contingent reward engineering to identifying and prioritizing questions where expert observations provide informative learning signals. Counterfactual probe attributes outcome variance to observations at the learner's own reasoning states, while PADS tracks these effects under the evolving policy and adaptively reallocates training data toward perception-dependent questions. This enables the policy to learn when to invoke experts through final-answer correctness alone, with only marginal additional rollout compute. Experiments and causal analysis confirm that data composition governs calling behavior: prioritizing observation-repairable questions increases Call Gain while reducing Call Harm, whereas imbalanced sampling induces excessive, insufficient, or spurious calls.



{
    \small
    \bibliographystyle{ieeenat_fullname}
    \bibliography{main}
}


\end{document}